\documentclass[10pt,twocolumn,letterpaper]{article}

\usepackage{cvpr} % uncomment this for the final submission
\usepackage{times}
\usepackage{epsfig}
\usepackage{graphicx}
\usepackage{amsmath}
\usepackage{amssymb}
\usepackage{booktabs}
\usepackage{threeparttable}
\usepackage{tikz}
\usepackage{float}

\usepackage[pagebackref=true,breaklinks=true,letterpaper=true,colorlinks,bookmarks=false]{hyperref}

\def\dag{\textsuperscript{\textdagger}}
\def\ast{\textsuperscript{*}}
\def\ddag{\textsuperscript{\textdaggerdbl}}
\def\sec{\textsuperscript{\S}}

\newcommand\copyrighttext{%
  \footnotesize \textcopyright 2026 IEEE. Personal use of this material is permitted. Permission from IEEE must be obtained for all other uses, in any current or future
  media, including reprinting/republishing this material for advertising or promotional purposes, creating new collective works, for resale or redistribution to servers or lists, or reuse of any copyrighted component of this work in other works.}
\newcommand\copyrightnotice{%
\begin{tikzpicture}[remember picture,overlay]
\node[anchor=south,yshift=10pt] at (current page.south) {\fbox{\parbox{\dimexpr\textwidth-\fboxsep-\fboxrule\relax}{\copyrighttext}}};
\end{tikzpicture}%
}

\begin{document}

%%%%%%%%% TITLE
\title{The 2$^{nd}$ International StepUP Competition for Biometric Footstep Recognition:\\ From Steps to Strides}

\author{Robyn Larracy\ast, Anant Gupta\dag, Gourav Gupta\dag, Ethan Eddy\ast, Maxime Devanne\ddag,\\ Cyril Meyer\ddag, Jin-Chern Chiou\sec, Yueh-Shan Lee\sec, Zong-Han Lu\sec, Aaron Tabor\ast, and Erik Scheme\ast \and
\ast University of New Brunswick, Canada\\ 
{\tt\small \{rlarracy,eeddy,aaron.tabor,escheme\}@unb.ca} \and
\dag ArogyaPandit Private Limited, India\\
{\tt\small \{anant.gupta,gourav.gupta\}@arogyapandit.com} \and
\ddag Université de Haute-Alsace, IRIMAS UR 7499, France\\
{\tt\small \{maxime.devanne,cyril.meyer\}@uha.fr}\and
\sec Institute of Electrical and Control Engineering, National Yang Ming Chiao Tung University, Taiwan\\
{\tt\small \{chiou,jojo.ee12,henrylu.ee11\}@nycu.edu.tw}}

\maketitle
\copyrightnotice
\thispagestyle{empty}

%%%%%%%%% ABSTRACT
\begin{abstract}

The International StepUP Competition Series was launched to advance research in pressure-based footstep biometrics through a standardized and challenging evaluation framework. 
Using the large-scale StepUP-P150 dataset (with more than 200,000 high-resolution dynamic footsteps from 150 individuals) and a previously unreleased test set, the 2nd edition of the competition addressed three key challenges: (1) generalization to unseen users with limited enrollment data, (2) robustness to domain shift caused by variations in footwear and walking speed and (3) effective fusion of paired left–right footsteps. 
While the first two challenges built on the inaugural competition, this edition introduced more extreme cross-domain conditions and moved beyond isolated footsteps to stride-level verification, enabling new opportunities for representation learning and inter-step information fusion. 
The competition attracted 26 registrants from academia and industry, with a best equal error rate of 8.00\% achieved by the ArogyaPandit Research Team using a spatiotemporal CNN combined with an ensemble-based scoring strategy. 
The top solutions showcase the value of harnessing temporal patterns and of incorporating inference-time normalization and calibration strategies to improve scoring. 
However, the results also reveal that recognizing users in unseen personal footwear remains a challenge, especially in the presence of distractors with similar characteristics.  

%By focusing on open-set verification under realistic variability, the competition aims to establish stronger baselines and accelerate progress in this emerging field.

\end{abstract}

%%%%%%%%% BODY TEXT
\section{Introduction}
% footstep recognition
As an emerging biometric modality, footstep recognition offers new possibilities for physical access control and activity monitoring in secure office buildings, airports, seaports, and other restricted environments.
Also called pressure-based gait recognition, these systems rely on underfoot pressure patterns recorded during walking using instrumented flooring or mats.
In addition to capturing gait patterns directly and unobtrusively from the foot's natural contact with the floor, this method of sensing can offer advantages over video-based alternatives due to its invariance to factors like lighting, occlusion, or camera positioning.

Despite growing interest in footstep biometrics, there are critical and underexplored challenges that need to be addressed to mature the technology for real-world deployment. 
Most notably, footstep patterns exhibit substantial intra-subject variability, arising from both intrinsic factors (e.g., natural behavioural and physiological variations) and extrinsic factors such as changes in footwear.
These covariates can cause significant domain shifts between the constrained and limited data captured during enrollment and the more diverse conditions encountered in practice. 
As a result, models must recognize users under conditions that differ substantially from their reference samples, while remaining robust to unauthorized users. 
Although promising verification and identification performance has been reported under controlled laboratory settings (e.g., 99.3\% identification accuracy for 55 barefoot participants \cite{Xie2020}, 98.5\% identification accuracy for 100 participants in familiar footwear \cite{Wei2026}), substantial performance degradation has been observed under variations in footwear, walking speed, and carried load \cite{Duncanson2023,Vera-Rodriguez2011,Moustakidis2008,Derlatka2017}. 
Addressing this gap remains a key barrier to advancing footstep recognition toward practical deployment.

The StepUP Competition series was launched to accelerate progress in this field through deep learning, powered by the University of New Brunswick's StepUP-P150 dataset: a resource of approximately 200,000 high-resolution footstep pressure maps from 150 individuals across a range of footwear types and walking speeds.     
The first iteration of the competition focused on single-footstep recognition, with a test protocol designed to assess generalization to unseen users and robustness to covariate shifts \cite{Larracy2025first}. 
Building on this foundation, the second edition introduces a newly designed test set with two key extensions. 
First, the probe set incorporates more extreme domain shifts, including barefoot samples and dynamically varying walking speeds. 
Second, rather than isolated footsteps, the probe set consists of pairs of consecutive left and right footsteps, providing additional context for recognition and enabling modelling at the inter-step and stride levels.
Together, these changes introduce more demanding evaluation conditions while simultaneously providing richer gait information for inference.
This paper presents the design and results of the second edition and identifies key methodological trends and open challenges to guide future research in footstep-based biometric recognition.

%This transition from individual steps to stride-based representations enables new approaches for representation learning, supporting the development of more robust and generalizable recognition models.
%By focusing on open-set verification under realistic variability, the competition aims to establish stronger baselines and accelerate progress in this emerging field.

\section{Datasets}

An overview of the training, reference, and probe datasets is provided in Table 1. 
All datasets are derived from high-resolution footstep recordings collected over an 18-month study conducted at the University of New Brunswick (REB 2022-132).
Each of the participants provided their informed consent for their participation in the study and the inclusion of their de-identified recordings and demographic information in a public repository.

\begin{figure}[b!]
    \centering
    \includegraphics[width=\linewidth]{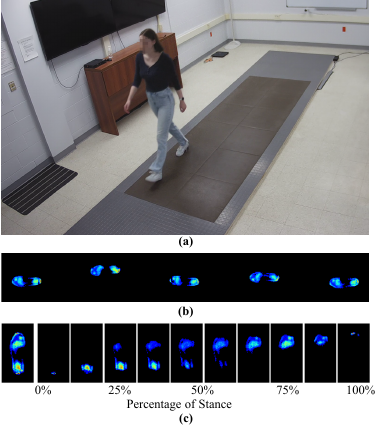}
    \caption{Study and dataset overview: (a) a participant walking across the instrumented walkway in a pair of personal shoes (flat canvas sneakers), (b) peak pressures captured from a sequence of five steps during one pass over the walkway, and (c) the progression of pressures from one pre-processed footstep recording, shown at ten different phases during the step.}
    \label{fig:example_step}
\end{figure}

Participants were instructed to walk back and forth across a 3.6 m \texttimes~1.2 m piezoresistive pressure-sensing platform (Stepscan Technologies Inc.), which captured high-resolution (4 sensors/cm$^2$) plantar pressure maps at a sampling rate of 100 Hz.
Full details of the collection protocol and data processing steps are provided in \cite{Larracy2025}; briefly, the protocol consisted of sixteen 90-second walking trials, collected with four different footwear conditions (BF: barefoot or sock-foot, ST: standard sneakers provided by the research team, and P1/P2: two pairs of personal shoes) and four different walking speeds (W1: preferred speed, W2: slowing to a stop, W3: slow, W4: fast).
The recordings were segmented into individual footsteps, which were spatially aligned and temporally normalized into tensors of size 101 $\times$ 75 $\times$ 40 (time, height, width).   
Using a multi-stage approach consisting of automated prediction and manual validation, the samples were annotated to indicate left and right steps, incomplete recordings, and outliers.    
Fig.~\ref{fig:example_step} shows an example of a participant walking across the instrumented tiles and the corresponding footstep recordings.

\begin{table}[t!]
\caption{Overview of the three datasets used for the competition. The reference and probe sets were comprised of samples from individuals not included in the training set.}
\setlength{\tabcolsep}{11pt}
\small
\begin{tabular}{@{}lccc@{}}
\toprule
\textbf{Dataset}   & \textbf{Users} & \textbf{Strides} & \textbf{Conditions}                           \\ \midrule
Training            & 150 & $>$ 100,000 & 4 speeds, 4 footwear \\
Reference & 15     & 75  & 1 speed, 1 footwear \\
Probe & 30           & 10,000             & 4 speeds, 4 footwear \\ \bottomrule
\end{tabular}
\end{table}

\subsection{Training Samples}
The training set comprises the complete collection of recorded footsteps from 150 participants (74 male and 76 female), with fully annotated data from all sixteen footwear–speed combinations. 
This dataset is publicly available as the UNB StepUP-P150 dataset for use by the gait and biometrics research communities \cite{Larracy2025} and comprises approximately 200,000 pre-processed footstep recordings.
Each recording is accompanied by rich metadata, including participant attributes (e.g., age, sex, height, weight, foot size, and personal footwear descriptions) and sample-level properties (foot laterality, rotation angle, position on the sensing grid, and walking direction). 
The cohort spans a broad range of physical characteristics and demographic backgrounds, with ages ranging from 19 to 91 years (mean 34 years), heights from 151 to 196 cm, body weights from 46 to 148 kg, and shoe sizes from UK 4 to 12.5.
Self-reported race/ethnicity categories include White ($N = 106$), Middle Eastern ($N = 15$), South Asian ($N = 10$), East/Southeast Asian ($N = 11$), Black ($N = 1$), multi-ethnic ($N = 5$), and unknown or unspecified ($N = 2$). 
In addition, the dataset captures a wide range of footwear choices, with 300 distinct pairs of personal shoes including sandals, athletic footwear, casual shoes, and boots.
These samples were made available to competitors as the primary resource for model development. 
The use of external data sources and pre-trained models was also permitted.

\subsection{Reference and Probe Samples}
The reference and probe sets comprised data from a separate set of 30 additional participants from the study that have never been made publicly available in an annotated form. 
The data from these participants were excluded from the StepUP-P150 dataset following collection (due to minor experimental deviations, incomplete recordings, or hardware issues, for example). 
The samples included in the reference and probe sets, however, underwent the same validation process as the training set to ensure label accuracy and recording quality. 
This cohort included 17 female and 13 male individuals with ages ranging 19-73 years and an average of 32.6 years.
Their heights ranged from 153 cm to 191 cm, weights from 42 to 147 kg, and shoe sizes from UK 3.5 to 12.5.
Self-identified racial groups included White ($N = 19$), Middle Eastern ($N = 3$), South Asian ($N = 3$), East/Southeast Asian ($N = 1$), Aboriginal ($N = 1$), multi-ethnic ($N = 2$), or unknown or unspecified race/ethnicity ($N = 1$).

%% reference samples
Consistent with the inaugural edition of the competition, fifteen participants from the test set were randomly selected as enrolled users to appear in the reference set. 
For each user, this reference set included five strides (i.e., five non-overlapping pairs of consecutive left/right steps) recorded from a single trial as the user walked in their personal footwear at their preferred speed.
The reference data were provided with only anonymized user identifiers, without any additional metadata related to participant characteristics or footwear. 
In total, the reference set consisted of 75 strides.
This deliberately limited reference set was designed to reflect realistic enrollment conditions, where only a small number of homogeneous samples may be acquired during a brief interaction (e.g., a few passes across the sensing platform). 
To ensure a novel evaluation setting, the identities of enrolled users were updated for this edition of the competition, and different footwear conditions were selected for inclusion in the reference set.

\begin{figure}[tb!]
    \centering
    \includegraphics[width=\linewidth]{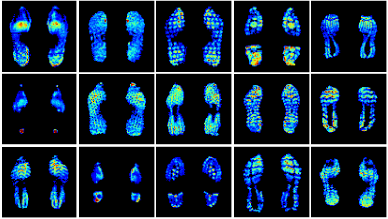}
    \caption{Examples of footstep pairs provided to competitors as labelled reference samples for each of the 15 enrolled users. Note that the strides are shown here as concatenated 2D pressure maps for visualization purposes; the competitors received two separate 3D (time, height, width) tensors for each footstep pair.}
    \label{fig:example_pairs}
\end{figure}

A challenging set of 10,000 probe strides was used for evaluation.
Each was randomly assigned a claimed identity associated with one of the enrolled users in the reference set,
resulting in a distribution of 25\% true claims and 75\% false claims. 
The true-claim probes included a small number of strides collected in known conditions (i.e., walking speeds and shoes included in the reference set), but the majority ($>90\%$) were captured under unseen conditions.
As noted, in addition to the slow and fast walking speeds and the other shod conditions (the standard shoes and the participants' second personal shoe type), this year's competition also incorporated the dynamic slow-to-stop and barefoot trials as unseen conditions in the probe set. 
These presented varying degrees of domain shift compared to the reference samples, making this a difficult generalization task.
The false-claim probes included strides from the 15 held-out distractor users and the other enrolled users.  
These also spanned all of the 16 footwear \texttimes~walking speed combinations.
As with the reference set, only the claimed identity was provided as metadata; ground truth labels were withheld.
Moreover, specific details regarding the composition of the probe set, including the proportions of the different conditions and the balance of true and false claims, were not disclosed to competitors. 
The public leaderboard and final rankings were based on the competitors' performance on this probe set, which was available in its entirety for the duration of the competition; no separate hidden final split was used for evaluation.

% not randomized uniformly: more likely for each shoe type to be assigned to one of the enrolled users. (less obvious for clustering).  
% approx equal proportions of three shoe types in both matched and non-matched
% more fast walking samples than slow or preferred

\subsection{Competition Task and Protocol} 
The competitors were challenged with verifying whether each of the 10,000 probe strides matched their claimed identity labels.
To compete, teams needed to submit two text files: one that contained a similarity score between 0 and 1 for each probe, and one that contained a single decision threshold used to convert the scores into match or non-match decisions. % one competitor submitted scores between -1 and 1. But it doesn't really change anything, and our scoring code handled it regardless

The competition was hosted using CodaBench \cite{codabench}, through which submissions were accepted from March to May 2026. 
The platform provided immediate performance feedback and allowed competitors to join a public leaderboard\footnote{\url{https:/codabench.org/competitions/13840}}.
Along with the three pre-processed footstep datasets, a Python-based code repository\footnote{\url{https:/github.com/UNB-StepUP/2nd_stepUP_competition}} was freely provided to competitors as a starting point. 
This repository included several utilities for data loading, visualization, and result formatting, as well as a baseline solution (see the details in Section \ref{sec:baseline}). 
% utilities for loading and preprocessing the three datasets in a memory-efficient way (i.e,, loaded during batch creation)
% grouping the training set into pairs to enable stride-level learning

Teams of up to six members were permitted. 
All teams were required to register in advance, providing team member names, affiliations, and a valid institutional or organizational contact email.
%To encourage iterative development while maintaining fairness, submissions were limited to five per day and up to fifty total per team. 
No restrictions were placed on computational resources, external data usage, or the use of pre-trained models. 
Submission of source code was generally not mandatory, to lower barriers to participation and accommodate those who may be subject to intellectual property or confidentiality constraints.
Thus, independent verification of submitted solutions was not performed, with the exception of accounts flagged for potential violations of the competition terms (see Section \ref{sec:participation}). 
Nevertheless, top-performing teams were encouraged to release their implementations publicly: where available, links to these repositories are provided along with the descriptions of methods in Section \ref{sec:solutions}.

To encourage iterative model development, competitors were permitted up to 50 submissions over the course of the competition, with a limit of five per day. 
Because aggregate performance feedback was provided for the probe set throughout the competition (i.e., in the form of summary error and accuracy rates), the evaluation did not rely on a strictly sequestered test set.
However, leaderboard overfitting was expected to be mitigated by the submission limit, the coarse nature of the provided feedback, and the scale and diversity of the 10,000-stride probe set.
Nonetheless, the reported error rates should be interpreted as measures of relative system performance rather than strict estimates of held-out generalization.

\subsection{Evaluation Metrics}
The submissions were ranked by their equal error rate (EER) on the probe set, representing the error rate at the decision threshold that best balances the false match rate (FMR) and false non-match rate (FNMR). 
Additional metrics were used for context and tie-breaking, including the FMR, FNMR, and balanced accuracy (BACC; computed as $100\% - (\mathrm{FMR} + \mathrm{FNMR}) / 2$), all evaluated at the team's submitted decision threshold. 
We also report FMR100, defined as the FNMR at a fixed FMR of 1\%.

\subsection{Baseline Solution}\label{sec:baseline}
An example PyTorch implementation was provided to competitors as a baseline solution. 
A spatiotemporal convolutional neural network (CNN) was used as the feature extraction backbone, with three blocks of factored spatial (2D) and temporal (1D) convolutions followed by adaptive max pooling to produce a 512-dimensional embedding for each footstep. 
The embeddings from the left and right footsteps in each stride were concatenated and fused using a fully connected layer to create the final vector of 512 features. % could add number of channels, kernel sizes as well

The network was trained end-to-end using a Circle Loss criterion \cite{Sun2020} with stride data from 140 training participants, with the remaining 10 participants reserved for validation. 
The Adam optimizer was used to train the model for 10 epochs with a batch size of 128 and an initial learning rate of $1\times10^{-4}$, decaying exponentially by a factor of 0.95 each epoch. 
Before training, footsteps were downsampled to 64 × 64 × 32 (time, height, width), amplitude values were normalized to the range [0,1], and right-foot samples were mirrored along the x-axis to improve training consistency.
For verification, the trained network was used to extract embeddings from each of the reference and probe footsteps, and a nearest-neighbour cosine similarity metric was used for scoring. 
This baseline solution achieved an EER of 14.12\% on the competition's probe set. 

\subsection{Participation}\label{sec:participation}
The competition was promoted through targeted emails to biometrics researchers and public announcements on LinkedIn and relevant Google Groups forums. 
In total, 26 teams registered for the competition, collectively submitting more than 600 entries, though the final standings reflect only the teams found to be in compliance with the competition terms. 
Participants represented both industry and academia, including undergraduate, Master’s, and PhD students, postdoctoral fellows, university faculty, and research staff. 
Competitors also reflected broad international participation, representing countries from Asia, Europe, North America, and Australia.

%%%%%%%%%%%%%%%%%%%%%%%%%%%%%%%%%%%%%%%%%%%%%%%%%%%%%%%%%%%%%%%%%%%%%%%%
\begin{table*}[tb]
\centering
\caption{Summary of the winning approaches.}
\small
\label{tab:approaches}
\setlength{\tabcolsep}{4pt}
\begin{threeparttable}
\begin{tabular}{p{2.5cm}p{2.0cm}p{1.4cm}p{2.2cm}p{2.3cm}p{1.6cm}p{3.5cm}}
\toprule
\textbf{Team} & \textbf{Backbone} & \textbf{Loss} & \textbf{Training \newline Strategies} & \textbf{L/R Fusion} & \textbf{Scoring} & \textbf{Inference-Time\newline Strategies} \\
\midrule
ArogyaPandit      & (2+1)D CNN  & Circle      & --                                          & Feature-level\newline (FC layer)                        & Prototype\newline cosine       & S-Norm (train cohort);\newline fusion of two score variants\\
ESquaredAnalytics & (2+1)D CNN \newline \& GRU    & Circle      & Data aug. (noise, shift, scaling,\newline masking) & Feature-level\newline (FC layer)                        & Prototype\newline cosine       & S-Norm (ref. cohort) \\
MC@MSD            & InceptionTime        & CE \& \newline SupCon      & Condition-aware \newline sampling                                          & Sensor-level\newline (channel concat.)            & 3NN cosine & BG score subtraction (ref. cohort) \\
Stride AI         & R(2+1)D CNN\newline \& 1D CNN & SupCon \& \newline ArcFace & Data aug. (hoz. \newline flip, crop, jitter)     & Sensor-level \newline (channel concat.) \newline \& feature-level \newline (cross-attn.) & Prototype \&\newline 3NN cosine & Four-view aug.; footwear \newline bias subtraction (train \newline cohort) \\
\midrule
Baseline          & (2+1)D CNN           & Circle      & --                                          & Feature-level\newline (FC layer)                        & NN cosine              & -- \\
\bottomrule
\end{tabular}
%\begin{tablenotes}
\footnotesize
CNN: Convolutional Neural Network; GRU: Gated Recurrent Unit; CE: Cross-entropy; FC: fully connected; NN: nearest neighbor; BG: background.
%\end{tablenotes}
\end{threeparttable}
\end{table*}
%%%%%%%%%%%%%%%%%%%%%%%%%%%%%%%%%%%%%%%%%%%%%%%%%%%%%%%%%%%%%%%%%%%%%%

Of the 26 registered teams, a handful of accounts were flagged based on patterns observed in the submission portal, including similarities in uploaded files and submission timing across accounts. 
Three of these flagged accounts had entries placing them within the top five teams; these competitors were contacted and asked to provide source code and model weights for verification. 
Following this process, in accordance with IEEE integrity standards, all three were formally disqualified for violating the independent participation policy by using multiple accounts to circumvent the per-team submission limit. 
%Although additional accounts exhibited similar patterns, their affiliation could not be established conclusively because they did not participate in the verification process. % remove?
The results reported in this paper reflect the leaderboard standings following the removal of disqualified accounts.

\section{Winning Solutions}\label{sec:solutions}

The organizers invited the winning teams to submit descriptions of their approaches. 
Methods from the top four teams are included here, and summarized in Table \ref{tab:approaches}.

\subsection*{ArogyaPandit Research Team (1st Place)} 
\noindent \textbf{Team}: Anant Gupta and Gourav Gupta, \textit{ArogyaPandit Private Limited, India}

\noindent \textbf{Code}: {\footnotesize \url{https://github.com/AnantG5/STEP-UP}}

Our final submission used an ensemble of two stride-level metric-learning systems trained only on the provided data. Participants 1-140 were used for training/cohort generation, and participants 141-150 were reserved for subject-disjoint validation. Each sample was represented as a paired left-right stride from the $64\times64\times32$ pressure sequence. We normalized pressure volumes to a common input shape and did not use external datasets or pretrained backbones.

The backbone is a Siamese 3D CNN shared between the left and right footsteps. The left and right foot embeddings are concatenated and passed through a fusion layer, then L2-normalized to obtain a 512-D stride embedding. We trained metric embeddings with Circle Loss ($\rm{m}=0.25$, $\rm{gamma}=256$), using validation EER for model selection. At inference, reference templates and probe strides were embedded with the same network. Claims were scored by cosine similarity against the claimed participant’s reference prototype, computed as the normalized mean of the available reference embeddings.

To reduce footwear/speed and cohort mismatch effects, we applied symmetric score normalization (S-norm) using training-set cohort embeddings. We generated S-norm variants using different cohort top-k values and reference aggregation rules. The submitted system fuses two complementary calibrated score lists: the strongest Circle-loss S-norm ensemble and a secondary balanced S-norm model. Because raw score ranges differ across models, we converted each list to empirical ranks and used weighted rank fusion: 88.5\% from the primary ensemble and 11.5\% from the complementary top-400 mean S-norm model. The final submitted scores.txt contained the fused claim scores and a fixed threshold file.

\subsection*{ESquaredAnalytics (2nd Place)} 
\noindent \textbf{Team}: Ethan Eddy, \textit{University of New Brunswick, Canada}

We extended the competition baseline along three axes: architecture, training, and inference.

For the architecture, we doubled the CNN backbone channels from [16, 32, 64] to [32, 64, 128] and appended a 2-layer GRU (hidden size 256, dropout 0.3) to model temporal dynamics across the footstep sequence. 
The CNN's spatial pooling was modified to preserve the temporal dimension as input to the GRU, whose final hidden state formed the per-footstep embedding.

For training, we added Gaussian noise, random temporal shifts, temporal scaling, pressure amplitude scaling in [0.5, 2.0], and random spatial masking of patches covering up to 30\% of the footstep dimensions. 
We extended training to 20 epochs, replaced exponential decay with cosine annealing, and raised the batch size to 256. 

At inference, we averaged each identity's reference embeddings into a single centroid rather than matching against the nearest reference stride. 
Cosine similarity scores were then normalized with cohort-based S-norm before computing EER.

\subsection*{MC@MSD (3rd Place)}
\noindent \textbf{Team}: Maxime Devanne and Cyril Meyer, \textit{Université de Haute-Alsace, IRIMAS UR 7499, France}

\noindent \textbf{Code}: {\footnotesize \url{https:/github.com/Cyril-Meyer/GaitUpStepUp}}

First, we preprocessed the original data. All left foot samples were flipped so that, in effect, only right foot patterns were used. We normalized pixel values using the logarithm of one plus the input, then divided the result by 7 to obtain values approximately in the [0,1] range.

Second, we trained a lightweight embedding extraction model on the training set. The model processes each video as a multivariate time series, where each footstep pixel is treated as a channel of length 101. This resulted in an input with 6000 channels, comprised of 3000 for each foot. We trained a single InceptionTime model using a combination of cross-entropy and contrastive loss applied on the embedding layer, the layer before the classifier. The loss is defined as $L = \rm{CE} + 0.2 \times \rm{ContrastiveLoss}$. For each batch, we sampled 4 subjects and 4 shoe/speed configurations. A batch was composed of 32 samples; 16 of which were drawn from the 4 selected shoe/speed conditions, and 16 of which were randomly drawn from any condition.
Third, we used the trained model to extract a 128-dimensional embedding vector for each footstep pair in the reference and probe sets. Finally, similarity scores were computed for each probe sample. We calculated a target similarity using the average cosine similarity between the probe embedding and the 3 nearest reference embeddings belonging to the claimed identity. We then computed a background similarity using the 5 nearest embeddings from different identities. The similarity score is defined as the difference between the target and background.

\subsection*{Stride AI (4th Place)} 
\noindent \textbf{Team}: Jin-Chern Chiou, Yueh-Shan Lee, and Zong-Han Lu \textit{Institute of Electrical and Control Engineering, National Yang Ming Chiao Tung University, Taiwan}

%\noindent \textbf{Task-informed Pipeline with Cross-Attention Fusion and Training-free Inference Calibration}

We developed the model using the 150 participants from the StepUP-P150 set across all four footwear (BF/ST/P1/P2) and four walking speed (W1/W2/W3/W4) conditions, using left–right footstep pairs as samples. 
Ten participants (IDs 141–150) were held out for validation, and the remaining 140 were used for training. 
No external datasets, pretrained backbones, or demographic metadata were used.

%\vspace{4pt}
%\noindent \textbf{Architecture:}
A lightweight R(2+1)D encoder (4 residual blocks) extracted left and right pressure embeddings (512-D), fused by a Cross-Attention Fusion Module modelling inter-foot dependencies. 
A 1D-CNN branch was used to process 6-channel COP and GRF features. 
Pressure (512-D) and COP/GRF (128-D) features were concatenated and projected to a 512-D embedding via a fully-connected fusion layer. 
Input tensors were resized to $32\times32\times32$.

%\vspace{4pt}
%\noindent \textbf{Training:}
For training, a joint loss combining SupCon (two-view, $\rm{temp}=0.07$) on a 128-D projection head and ArcFace ($s=16$, $m=0.5$, $\rm{weight}=0.05$) on the 512-D embedding was used. 
The Adam optimizer ($\rm{lr}=3e-4$, ExponentialLR $\gamma=0.98$) was used during training with a batch of 32 identities × 4 samples for 50 epochs. 
For data augmentation, horizontal flipping ($p=0.3$), temporal cropping (0.8) and pressure-scale jitter were applied. 
Pressure tensors were min-max normalized.

%\vspace{4pt}
%\noindent \textbf{Inference (Training-Free):}
During inference, four-view test-time augmentations (original, h-flip, temporal-crop, both) were used to create averaged L2-normalized embeddings. 
We then applied Symmetric Global Bias Subtraction: a footwear-mean vector was precomputed for each of the four shoe classes over all 150 training participants. 
As true footwear labels were unavailable at test time, the shoe class of each sample was estimated label-free via a mean-pressure threshold; the corresponding footwear mean was then subtracted from each embedding with weight $\alpha=2.5$, projecting both sides into a footwear-neutral space. 
The final score fused prototype cosine (weight 0.6) with top-3 nearest-neighbour cosine (weight 0.4), followed by min-max calibration against the validation score distribution.
Additional details about the architecture and inference pipelines are available in the Supplementary Material (see Figures S1 and S2). 

\section{Results}
%%%%%%%%%%%%%%%%%%%%%%%%%%%%%%%%%%%%%%%%%%%%%%%%%%%%%%%%%%%%%%%%%%%%%%%%%%%%%%%%%%%%%%%%%%%%%%
% Tables & figs
%%%%%%%%%%%%%%%%%%%%%%%%%%%%%%%%%%%%%%%%%%%%%%%%%%%%%%%%%%%%%%%%%%%%%%%%%%%%%%%%%%%%%%%%%%%%%%
% leaderboard (add AUC?)
\begin{table}[tb!]
\small
\setlength{\tabcolsep}{3.1pt}
\caption{Performance metrics (in \%) for the top four ranked teams and the baseline model. EER and FMR100 were computed using optimized decision thresholds, and the remaining metrics (BACC, FNMR, FMR) were computed at the decision threshold submitted by the team.}
\begin{tabular}{@{}llcccccc@{}}
\toprule
\# & Team Name & EER & FMR100 & BACC & FNMR & FMR \\ \midrule
1 & ArogyaPandit   & 8.00 & 42.88  & 83.52 & 0.16  & 32.80 \\
2 & ESquaredAnalytics      & 8.05 & 52.99 & 86.57 & 0.88  & 25.97 \\
3 & MC@MSD  & 8.88 & 48.24 & 89.97  & 14.88 & 5.17 \\
4 & Stride AI & 10.27 & 62.63 & 89.20 & 6.24 & 15.36 \\
\midrule
  & Baseline & 14.12 & 49.68 & 82.83 & 29.84 & 4.49 \\
\bottomrule
\end{tabular}
\label{tab:leaderboard}
\end{table}

\begin{figure}[tb]
    \centering
    \includegraphics[width=\linewidth]{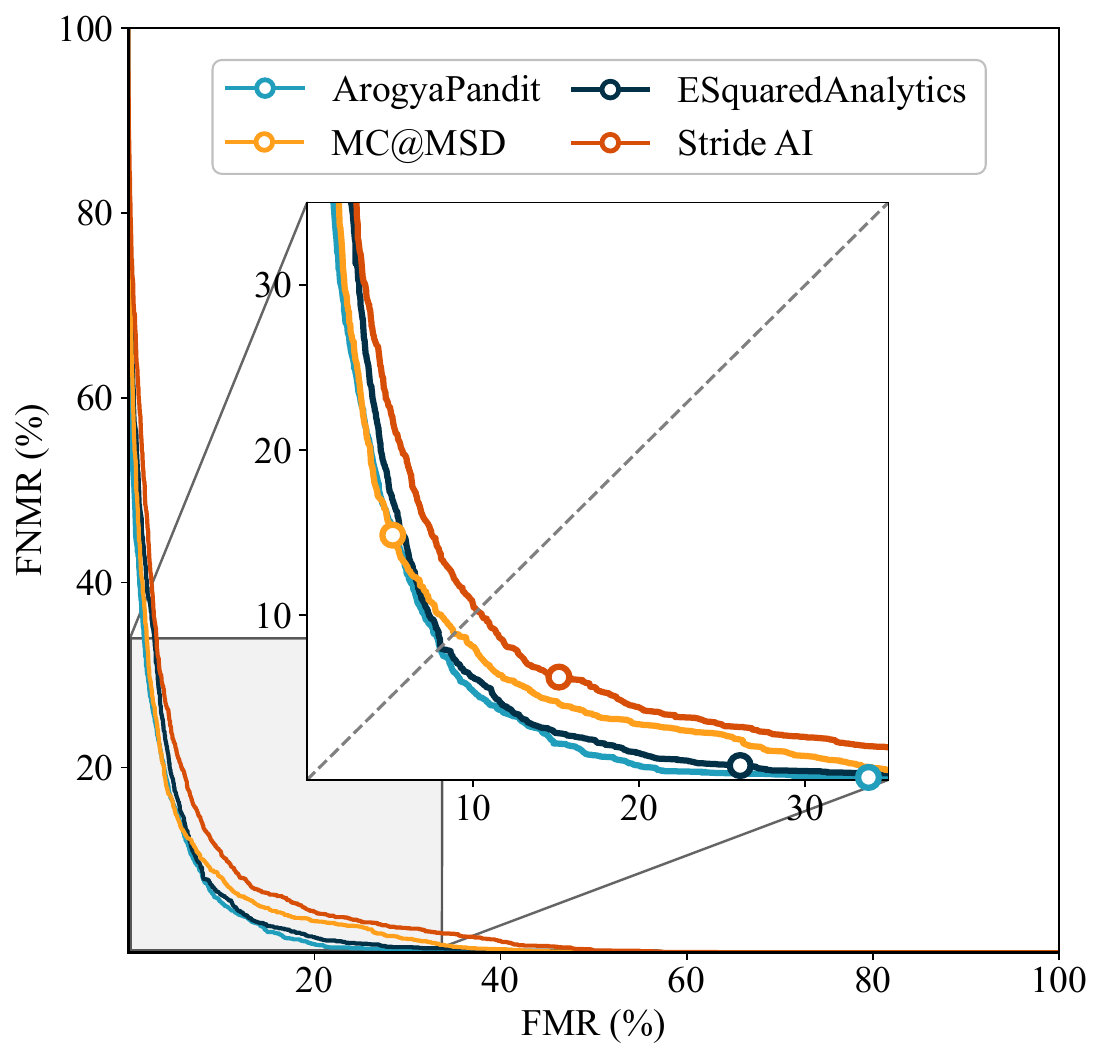}
    \caption{Detection error trade-off (DET) curves for the top four teams. The operating points submitted by the competitors are indicated with circle markers.}
    \label{fig:det_curves}
\end{figure}

%%%%%%%%%%%%%%%%%%%%%%%%%%%%%%%%%%%%%%%%%%%%%%%%%%%%%%%%%%%%%%%%%%%%%%%%%%%%%%%%%%%%%%%%%%%%%%

Table \ref{tab:leaderboard} summarizes the probe set performance of the top four teams, ranked by equal error rate (EER). The ArogyaPandit Research Team achieved the best overall performance with an EER of 8.00\%, followed closely by ESquaredAnalytics at 8.05\%. 
Although ranked third overall by EER, the MC@MSD team achieved the highest balanced accuracy (89.97\%), largely due to a more conservative threshold selection that resulted in a lower false match rate (FMR) than the other submissions. 
Figure \ref{fig:det_curves} further illustrates the error trade-off characteristics of the top-performing approaches, all of which exhibited a tendency toward convenience (low FNMR) rather than security (low FMR).

\begin{figure}[t]
\centering
\includegraphics[width=\linewidth]{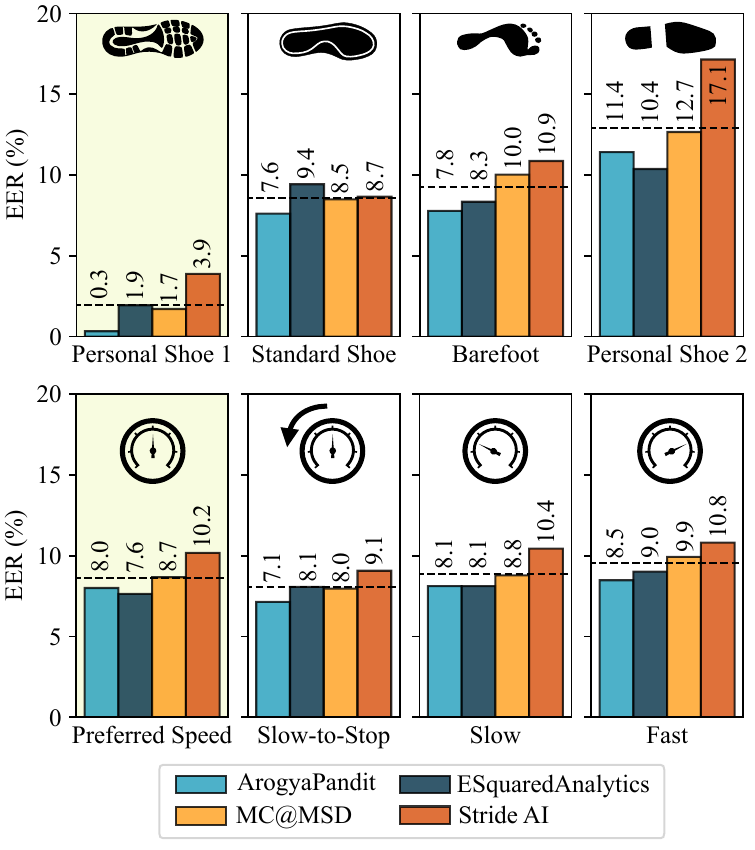}
\caption{Probe set EERs for each team, stratified by footwear conditions (top) and walking speeds (bottom). The conditions in the first column (Personal Shoe 1, Preferred Speed) are the same as those in the reference set, while the other conditions represent domain shifts between enrollment and verification. Averages are indicated with black dotted lines.}
\label{fig:EER_by_condition}
\end{figure}

Figure \ref{fig:EER_by_condition} presents each team's performance stratified by footwear and walking speed, which includes EERs computed for conditions represented in the reference set (Personal Shoe 1 and Preferred Speed) as well as previously unseen conditions for each user. 
Generalization across footwear conditions proved challenging for all teams, with average EER increases of 6.6\%, 7.3\%, and 10.9\% for the standard shoe, barefoot, and alternate personal footwear conditions, respectively. 
In contrast, changes in walking speed had minimal impact on performance, indicating that the submitted models were largely robust to variations in gait velocity.

%%%%%%%%%%%%%%%%%%%%%%%%%%%%%%%%%%%%%%%%%%%%%%%%

\begin{figure}[tb]
    \centering
    \includegraphics[width=\linewidth]{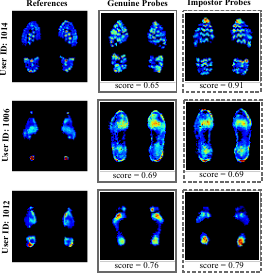}
    \caption{Examples of three challenging cases in the probe set, where false-claim (impostor) probes had the same or higher similarity scores than true-claim (genuine) probes from the enrolled user. Similarity scores from the top team's solution are shown, although similar trends were observed across submissions.}
    \label{fig:challenging_probes}
\end{figure}
%%%%%%%%%%%%%%%%%%%%%%%%%%%%%%%%%%%%%%%%%%%%%%%%

There were several particularly difficult cases in the probe set that deceived all four of the top submissions. 
These were identified by averaging the rank-transformed scores from the top four teams, and surfacing the users with the most score overlap between true-claim (genuine) and false-claim (impostor) comparisons. 
% genuine probes that scored unexpectedly low and impostor probes that scored unexpectedly high across all systems.
Examples of challenging probes are shown in Fig. \ref{fig:challenging_probes}; all cases included footwear conditions that differed from those used during enrollment. 
Some of these presented very large changes in both gait mechanics and the soles' interface with the sensors, such as comparing references in high heels with claimants in standard shoes (casual Adidas sneakers), as shown in the second row.

Under these challenging conditions, impostor probes frequently produced the same or higher similarity scores than corresponding genuine probes. 
This was observed especially for distractors with similar foot sizes and/or footwear characteristics to the claimed enrollee. 
In one case (bottom row of Fig. \ref{fig:challenging_probes}), a distractor with a pronounced foot arch had a similar impression to the high-arch business shoe that the enrolled user wore during enrollment -- these samples were often scored higher than genuine barefoot samples from the enrolled user, which did not have this high-arch characteristic.
Notably, other factors such as age and weight did not appear to strongly influence probe difficulty. 
For example, the users shown in the bottom row of Fig. \ref{fig:challenging_probes} differed in weight by more than 20 kg, while age differences across the three examples ranged from 17 to 48 years. % added a quick discussion point suggesting use of soft-biometric information to improve rejection

%Of the 10,000 probes, there were 223 strides that were misclassified by all of the top teams at their provided decision thresholds; all of these were false matches. 
% to identify the difficult cases above, decided to focus on raw scores/their overlap rather than performance at the competitor's thresholds since they were a bit poorly selected this year

\section{Discussion}

% domain generalization
Similar to the first edition of the competition, the solutions demonstrated that performance for footwear conditions represented in the training set (standard shoe and barefoot) was substantially better than performance when generalizing to arbitrary unseen personal footwear styles. 
Likewise, little variation in performance was observed across the four walking speed conditions, which were also represented in the training data for all 150 participants. 
These findings suggest that the models were able to learn footwear- and speed-related mappings that generalized effectively to a separate population of users under similar conditions.
However, complete domain generalization, that is, the recognition of new users wearing shoe styles that the model has never seen during training, remains an open challenge. 
It is possible that greater footwear diversity, achieved through additional data collection or simulated training data, could enable models to learn more robust shoe-invariant representations. 
Despite the size and diversity of the StepUP dataset, several highly specific footwear domains remain underrepresented, such as high heels and flip-flops.

% score distribution
The model scores for all four of the top solutions generally favoured convenience (low FNMR) over security (low FMR), as reflected in the shapes of the DET curves. 
The best performance at a high-security operating point (FMR of 1\%) was achieved by the ArogyaPandit Research Team, with an FNMR of 42.88\%. 
This asymmetry likely reflects the distribution of the probe set, in which the vast majority of genuine probes represented previously unseen conditions, encouraging more permissive system behaviour at both the feature extraction and scoring stages. 
Bridging these domain gaps and handling the high intra-class variability required models to favour lower rejection rates under challenging conditions.
Consequently, the models were encouraged to focus on domain-invariant rather than domain-specific cues, such as shoe sole patterns or foot shape, which may improve rejection performance under constrained conditions but generalize poorly across domains.
Future work should investigate shoe-aware or condition-aware approaches that leverage domain information when appropriate without compromising generalization or security. % i.e., identify the shoe type at inference time and use that knowledge to condition the model decisions
Moreover, because the models occasionally struggled to reject distractors despite clear differences in factors such as age and body weight (e.g., the $>20$ kg weight difference between users in the last row of Fig. \ref{fig:challenging_probes}), future work could explore estimating these attributes from the footsteps \cite{Chen2021,Zhang2025} and incorporating them as soft biometric traits to improve rejection performance.
%These findings also highlight the importance of threshold selection and operating-point analysis for safety-critical applications involving high variability and domain shift.

% treatment of multiple footsteps (fusion strategies)
The use of footstep pairs rather than individual footsteps led to improved overall performance in this year’s competition, despite the inclusion of additional domain shifts in the test set. 
Whereas the top five EERs in the inaugural competition ranged from 10.77\% to 12.23\%, the best-performing approaches in this edition achieved EERs ranging from 8.00\% to 10.27\%. 
This improvement likely reflects both the complementary information provided by the two sides, given the natural asymmetries in human gait \cite{Sadeghi2000}, and the increased robustness gained by aggregating evidence across consecutive measurements.
The four top teams took notably different approaches to fusing left and right footsteps. 
The top two teams followed the provided baseline and used feature-level fusion, employing a shared backbone and a single fully connected layer to merge embeddings from the two sides. 
The MC@MSD team instead opted for early fusion, concatenating the left and right pixel time series channel-wise before passing the combined input to their InceptionTime network. 
The Stride AI team incorporated two strategies: an early fusion approach, concatenating ground reaction force and center of pressure time series from both sides as input to a 1D CNN, and a feature-level approach with a cross-attention module to merge left and right embeddings from a shared R(2+1)D encoder. 
Although early fusion through concatenation of preprocessed left and right representations has been the dominant strategy in prior footstep recognition literature \cite{Iskandar2025,Horst2023,Duncanson2023,Wu2025,CostillaReyes2018}, there is still no clear consensus on the most effective fusion strategy.
Consequently, future work should investigate the impact of fusion at different stages of the recognition pipeline, as well as the incorporation of longer sequences of consecutive footsteps where sensor configurations and environments permit.
% could potentially talk about incorporating other stride-level features; stride length, angles, etc. But these features were not available to the competitors for the probe set.

% trends in approaches: score normalization
Of note, a theme among this year's solutions was the incorporation of score normalization and other inference-time strategies.
Although the baseline did not include score normalization, the top three teams adopted cohort-based approaches to improve scoring under the substantial domain shift; two solutions used the other reference samples (14 other users) as the cohort, while the top team (ArogyaPandit Research) incorporated data from the training set (140 users).
These approaches yielded substantial gains over the baseline despite relatively modest architectural changes, as with the ESquaredAnalytics and ArogyaPandit Research solutions. 
Similarly, the Stride AI team introduced several inference-stage enhancements, including test-time augmentation and subtraction of footwear-specific feature vectors derived from the training set.
Inference-time normalization or calibration strategies are a promising avenue for further exploration, with the expense of additional processing requirements during transactions. 
Moreover, models must be evaluated without information leakage across test probes to accurately reflect performance in single-presentation, real-world use cases. 
%It should be noted that the top solution also employed a score fusion strategy that used knowledge of other probes for normalization; while this was not prohibited in the competition, this is a procedure that would not be applicable in practical deployment scenarios where probes are processed individually and the full score distribution is not available at inference time.

% trends in approaches: temporal modeling
Another recurring theme among this year's top submissions was a focus on explicitly leveraging temporal patterns in the pressure sequence.
While the baseline used a spatiotemporal CNN as the feature extraction backbone, three of the four top teams incorporated architectures or representations more directly tailored to temporal modeling: two used architectures native to time series classification (GRUs and InceptionTime), and one extracted ground reaction force and center of pressure time series (representations previously shown to be less sensitive to footwear variability than spatial pressure maps \cite{Kulkarni2024}) as input to a 1D CNN. 
Notably, the InceptionTime-based solution from the MC@MSD team modelled each pixel's time series as an independent channel, forgoing explicit spatial modelling entirely while still achieving competitive performance.
InceptionTime has previously shown strong robustness for recognition from force plate recordings under asymmetric loading \cite{Derlatka2026}. 
This year's approaches are in contrast to the first edition of the competition where the top three teams all relied on R(2+1)D CNNs for spatiotemporal feature extraction \cite{Larracy2025first}, although this architecture was incorporated as a component of the Stride AI solution.
Collectively, these results suggest that the temporal evolution of the footstep pressures carries robust discriminative information, motivating further exploration of time-series-oriented architectures for this modality.
% none of the solutions opted to use pretrained models or foundation models; given the specialized sensing modality, these may not transfer well. 

% opportunities to use soft biometric information, metadata to enhance training
% shoe subtraction approach

\section{Conclusions}
This paper presented the results of the second edition of the StepUP Competition, which extended the evaluation protocol to consecutive footstep pairs while introducing additional domain shifts in the probe set.
The use of paired left-right footsteps contributed to improved recognition performance relative to the first edition, with top solutions establishing strong baselines for this challenging evaluation setting. 
The best-performing approaches demonstrate the value of both well-designed deep learning architectures that explicitly leverage temporal patterns in pressure sequences, and inference-time techniques such as score normalization and ensembling strategies. 
Although challenging cases remain, particularly involving unfamiliar and rare footwear styles, the overall performance trends suggest that cross-footwear and cross-speed domain shifts are learnable and can generalize to unseen users.
Future work should also investigate the impact of demographic factors on recognition performance, as certain groups remain underrepresented in the dataset and subgroup-level evaluation has yet to be systematically explored.
As in the first edition, the competition platform will remain open on CodaBench, and the organizers encourage further submissions from the research community to benchmark new algorithms against this dataset and protocol.

\section*{Acknowledgments}
This project was supported by the New Brunswick Innovation Foundation's Strategic Opportunities Fund, the Atlantic Canada Opportunities Agency's Regional Innovation Ecosystem program, and the Natural Sciences and Engineering Research Council of Canada's Alliance Grants program [ALLRP 558340-20] and Postgraduate Scholarship - Doctoral award [PGS D-589922-2024].

The organizers disclose that the 2\textsuperscript{nd} place team is affiliated with the host laboratory. However, the competitor operates within a distinct research domain and entered the competition independently. They were granted no privileged discussion, preferential access to challenge data, or institutional resources. The competition maintained a strict blind evaluation process, subjecting all participants to uniform rules and submission guidelines.

The organizers express their gratitude to all teams for their participation and contributions.

{\small
\bibliographystyle{ieee}
\bibliography{egbib}
}

\pagebreak
\onecolumn
\renewcommand{\thefigure}{S\arabic{figure}}
\setcounter{figure}{0}

\section*{Supplementary Material}
\subsection*{Additional Detail for Stride AI Solution}

\begin{figure}[h!]
    \centering
    \includegraphics[width=0.5\linewidth]{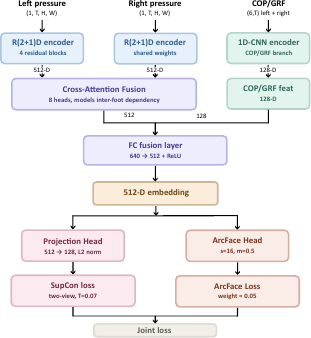}
    \caption{Training architecture.}
    \label{sfig:training_architecture}

    \vspace{10pt} 
    Key components shown in the diagram:
    \begin{itemize}
    \item Left and right pressure tensors are each fed into a shared-weight R(2+1)D encoder
    \item A separate 1D-CNN branch processes the 6-channel COP/GRF features
    \item Left and right pressure features attend to each other via a Cross-Attention Fusion module
    \item The fused pressure features are concatenated with the COP/GRF features and passed through an FC fusion layer to produce a 512-D embedding
    \item The same embedding feeds two loss heads: SupCon (through a projection head down to 128-D) and ArcFace (directly on the 512-D embedding)
    \end{itemize}
\end{figure}

\begin{figure}[h!]
    \centering
    \includegraphics[width=0.5\linewidth]{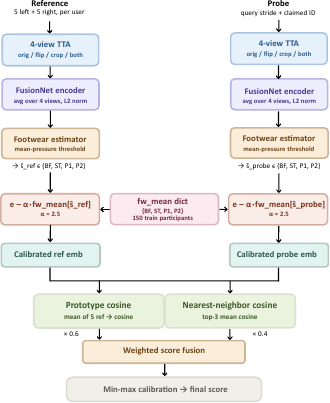}
    \caption{Inference pipeline.}
    \label{sfig:inference_pipeline}

    \vspace{10pt}
    Key components shown in the diagram:
    \begin{itemize}
    \item Reference samples (5 left + 5 right strides per enrolled user) and the probe stride each pass through 4-view Test-Time Augmentation(original / horizontal-flip / temporal-crop / both).
    \item Each set of views is fed through the trained FusionNet encoder, then averaged and L2-normalized to obtain a single embedding.
    \item A footwear-mean vector is precomputed for each of the four shoe classes (BF/ST/P1/P2) over all 150 training participants.
    \item As true footwear labels are unavailable at test time, a label-free mean-pressure threshold estimates the shoe class of each reference and probe sample.
    \item Symmetric Global Bias Subtraction: the corresponding footwear-mean is subtracted from each embedding with weight $\alpha=2.5$, projecting both sides into a footwear-neutral space.
    \item Two scores are computed in parallel: Prototype cosine (probe vs. mean of 5 references) and Nearest-Neighbor cosine (mean of top-3 cosines).
    \item Weighted score fusion combines the two scores with 0.6 × prototype + 0.4 × NN, followed by min-max calibration against the validation score distribution to produce the final verification score.
    
    \end{itemize}
\end{figure}

\end{document}